\documentclass{article}

\PassOptionsToPackage{numbers, compress}{natbib}

\usepackage{caption}

\usepackage{amsmath}
\usepackage{makecell}
\usepackage{subcaption}
\usepackage{multirow}
\usepackage{longtable}
\usepackage[preprint]{neurips_2026}

\usepackage[utf8]{inputenc} 
\usepackage[T1]{fontenc}    
\usepackage{hyperref}       
\usepackage{url}            
\usepackage{booktabs}       
\usepackage{amsfonts}       
\usepackage{nicefrac}       
\usepackage{microtype}      
\usepackage{xcolor}         
\usepackage{graphicx}
\title{When Does Task-Aware Imputation Matter? An Analysis of Robustness Under Train–Test Mismatch}

\author{Maria V.~Salvador\\
  INESC TEC\\
  University of Porto\\
  \texttt{maria.v.salvador@inesctec.pt} \\
\And
Ana F. Sequeira \\
INESC TEC \\
University of Porto \\
\texttt{ana.f.sequeira@inesctec.pt}
\And
Jaime S Cardoso \\
INESC TEC \\
University of Porto \\
\texttt{jaime.cardoso@inesctec.pt}
}

\begin{document}

\maketitle

\begin{abstract}
Missing data is a pervasive challenge in machine learning, often degrading predictive performance and complicating model deployment. Most existing approaches treat imputation as a preprocessing step independent of the downstream task, a separation that becomes problematic under train–test mismatches in missingness mechanisms or rates. We introduce a unified framework that treats missing entries as variables in a task-conditioned optimization problem. We instantiate this framework in two settings: an amortized approach for scenarios with missing data at both training and test time, and a latent optimization approach for settings where only training data is incomplete.
The evaluation is conducted on 98 tabular datasets and robustness under train–test mismatches is assessed on 36 datasets.
Results reveal that reconstruction error constitutes a poor proxy for predictive utility (Spearman $\rho\approx$ 0.16–0.32), and that task-aware imputation becomes critical in scenarios where classical methods fail to recover complete-data performance, achieving win rates of up to 80\% in challenging train–test mismatch scenarios.

\end{abstract}

\section{Introduction}
Missing data is a pervasive problem in real-world machine learning, with significant implications for the validity, reliability, and fairness of predictive models. In practice, data may be incomplete due to collection errors, equipment failures, participant drop-out, or reluctance to disclose sensitive information. Missing values can lead to biased estimates, reduced statistical power, and degraded performance, making their handling a critical component of any data-driven pipeline.

Understanding why data is missing-- the missingness mechanism-- is essential. There are three main missingness mechanisms: Missing Completely At Random (MCAR), where the probability of missingness is independent of the data; Missing At Random (MAR), where the probability of missingness is independent of the missing values given the observed data; and Missing Not At Random (MNAR), where the probability of missingness depends on the unobserved values themselves \cite{Kleinke2020,little2019}. These mechanisms not only influence how missing data should be modeled, but also interact with when missingness occurs in the learning pipeline.




In practice, an important consideration is whether missingness occurs at inference time. In some applications, missing values appear only during training, while deployment data is clean \cite{little2019}. In others, missingness persists at test time, often under shifts in mechanisms or rates \cite{DBLP:conf/aistats/ZhouBL23,10446980}. These scenarios pose distinct challenges: the former primarily affects representation learning, while the latter requires robustness to distribution shift. Despite these differences, most imputation methods treat missing data as a static preprocessing problem, without explicitly accounting for when or how missingness occurs.

This raises a fundamental question: to what extent does improving imputation quality translate into better predictive performance, particularly in realistic settings where missingness persists at test time or differs between training and deployment?
This question leads to the primary goal of this work, which is to investigate when task-aware imputation improves predictive performance in supervised learning with missing values.

Our contributions are threefold.
First, we show that task-aware imputation is not universally beneficial, but instead becomes critical in challenging regimes such as distribution shift and intrinsically hard missingness settings.
Second, we introduce a simple and flexible formulation that treats missing values as learnable variables jointly optimized with the predictive model, providing a common perspective that encompasses both amortized and latent approaches within a single framework.
Third, through a comprehensive empirical study, we demonstrate that standard reconstruction metrics are only weakly correlated with downstream performance, highlighting the limitations of conventional evaluation protocols.

\section{Related Work}
Approaches to handling missing data include imputation, where missing entries are filled; omission, where incomplete samples are discarded; and methods that directly operate on incomplete data \cite{little2019,LI201415,rubin}. In this work, we focus on imputation.

Classical methods such as mean, median, and $K$-NN \cite{Batista01052003} are simple and efficient but fail to capture complex dependencies. Iterative approaches such as MissForest \cite{10.1093/bioinformatics/btr597}, MICE \cite{mice}, and related conditional models \cite{albu2025missforestpredict} model non-linear relationships but are computationally expensive and sensitive to design choices. HyperImpute \cite{Jarrett2022HyperImpute} extends this paradigm through automated model and hyperparameter selection, while low-rank matrix completion methods exploit global structure \cite{candes2009exact, mazumder2010spectral}. Learning-based approaches such as TabPFN \cite{hollmann2025tabpfn} and TabImpute \cite{feitelberg2025tabimputeaccuratefastzeroshot} leverage pretrained models for fast, zero-shot imputation. Deep learning methods for tabular data, including transformer-based architectures \cite{gorishniy2021revisiting,10.5555/3524938.3525599}, have further advanced this direction. 

Several approaches model the joint data distribution using deep generative models \cite{du2024remasker,10.1007/978-3-030-01768-2_8,vincent2008extracting,Kuang2025UnsupervisedDI,mattei2019miwae,nazabal2020hivae,yoon2018gain,10.5555/3737916.3741474,jolicoeurmartineau2023generating}, learning to reconstruct missing entries through reconstruction objectives, often with masking strategies during training. While expressive, these methods typically prioritize data fidelity over downstream predictive performance. 
Other works incorporate additional structure into the imputation process. For instance, MIRACLE \cite{kyono2021miracle} models the missingness mechanism under MNAR using causal assumptions, while optimal transport-based approaches align distributions between observed and missing data \cite{leeUnbalancedOptimalTransport2020,peyreComputationalOptimalTransport2019}.

Handling missing data in supervised learning has traditionally relied on two-stage pipelines, where missing values are first imputed and then used for downstream prediction. This separation can lead to suboptimal performance, as imputation is not aligned with the predictive task. Benchmark studies show that the choice of imputation method significantly affects downstream performance, particularly in terms of robustness and interpretability \cite{shadbahr2023imputation}. Moreover, standard metrics such as RMSE and MAE often fail to reflect imputation quality, as they do not capture discrepancies in the underlying data distribution.

Task-aware approaches aim to address this limitation. TARImpute \cite{10.14778/3750601.3750667} co-optimizes imputation with clustering objectives, while OptImpute \cite{bertsimas2018} formulates imputation as an optimization problem independent of downstream supervision. Deep learning methods such as GRAPE \cite{you2020handling} and GEDI \cite{chen2023gedi} propose end-to-end frameworks that jointly perform imputation and prediction, often relying on specific architectural choices such as graph-based representations.

Changes in missingness mechanisms can also induce distribution shift. Domain Adaptation under Missingness Shift (DAMS) \cite{DBLP:conf/aistats/ZhouBL23} shows that differences in missingness alone can invalidate standard assumptions and require adaptation, particularly when missingness indicators are unavailable. Similarly, StableMiss+ \cite{10446980} studies mask distribution shifts and proposes removing unstable correlations involving the missingness mask.

These works highlight two key challenges: (i) robustness under shifts in missingness mechanisms or rates, and (ii) the misalignment between imputation quality and downstream predictive performance.
We address these challenges by introducing a unified formulation that treats missing values as learnable variables jointly trained with the predictive model. This perspective directly aligns imputation with the task objective, and allows us to characterize when such alignment translates into improved predictive performance.
Unlike prior task-aware approaches, our formulation does not depend on specific architectures or specialized designs (e.g., graph-based models), and avoids the need for bilevel optimization. Instead, it provides a simple and flexible lens to study task-aware imputation itself, encompassing both per-sample (latent) and amortized strategies within a single formulation. 
Furthermore, we explicitly investigate train--test mismatches in missingness mechanisms and rates, a setting that remains largely underexplored in existing work.

\section{Joint-AmortizedImpute: Task-Aware Imputation via Joint Optimization}

Let $\mathcal{D} = \{(\mathbf{x}_i, y_i)\}_{i=1}^n$ be a dataset of $n$ samples, where:
\begin{itemize}
  \item $\mathbf{x}_i \in \mathbb{R}^d$ is the input feature vector for the $i$-th sample,
  \item $y_i \in \mathcal{Y}$ is the corresponding label or target value,
  \item some components of $\mathbf{x}_i$ may be missing.
\end{itemize}

Let $M \in \{0,1\}^{n \times d}$ be a binary mask matrix such that:
\[
m_{ij} = 
\begin{cases}
1 & \text{if } x_{ij} \text{ is observed}, \\
0 & \text{if } x_{ij} \text{ is missing}.
\end{cases}
\]

The learning objective is to find a function $f_\theta: \mathbb{R}^d \rightarrow \mathcal{Y}$, parameterized by $\theta$, that minimizes the expected loss:

\begin{equation}
    \min_{\theta} \mathbb{E}_{(\mathbf{x}, y) \sim \mathcal{P}} \left[ \ell(f_\theta(\mathbf{x}), y) \right],
\end{equation}

where $\ell$ is a loss function and $\mathcal{P}$ is the underlying data distribution.

However, due to missing values, we only observe $\mathbf{x}_i^{\text{obs}} = \mathbf{x}_i \odot \mathbf{m}_i$, and do not observe  \\ 
$\mathbf{x}_i^{\text{u}} = \mathbf{x}_i \odot (\mathbf{1}-\mathbf{m}_i)$,
where $\odot$ denotes element-wise multiplication. 
The learning problem becomes:
\begin{equation}
    \min_{\theta} \mathbb{E}_{(\mathbf{x}^{\text{obs}}, y) \sim \mathcal{P}_{\text{obs}}} \left[ \ell(f_\theta(\mathbf{x}^{\text{obs}}, \mathbf{\hat{x}}^{\text{u}}), y) \right],
\end{equation}
where $\mathbf{\hat{x}}^{\text{u}}$ is an estimate of $\mathbf{x}^{\text{u}}$ 
with an imputation strategy.

We propose to jointly optimize the predictive model and the missing values in a unified objective. Let $\mathbf{x}^{\text{obs}}$ and $\mathbf{x}^{\text{u}}$ denote the observed and unobserved components of the input, respectively. We consider the following objective:
\begin{equation}
\min_{\theta, \mathbf{x}^{\text{u}}} 
\mathbb{E}_{(\mathbf{x}^{\text{obs}}, y) \sim \mathcal{P}_{\text{obs}}} 
\left[ \ell\big(f_\theta(\mathbf{x}^{\text{obs}}, \mathbf{x}^{\text{u}}), y\big) \right].
\end{equation}

To regularize the solution, we introduce a penalty on the unknown components:
\begin{equation}
\min_{\theta, \mathbf{x}^{\text{u}}} 
\mathbb{E}_{(\mathbf{x}^{\text{obs}}, y)} 
\left[ \ell(f_\theta(\mathbf{x}^{\text{obs}}, \mathbf{x}^{\text{u}}), y) \right] 
+ g(\mathbf{x}^{\text{u}}),
\end{equation}
where $g$ encodes prior assumptions about plausible imputations.

\paragraph{Reference-based parameterization.} We express missing values as deviations from a reference imputation $\hat{\mathbf{x}}^{\text{u}}$:
\begin{equation}
\mathbf{x}^{\text{u}} = \hat{\mathbf{x}}^{\text{u}} + \boldsymbol{\delta},
\end{equation}

This parameterization can be interpreted as introducing a strong prior over plausible imputations through the reference solution. It stabilizes optimization by anchoring the search in a realistic region of the input space, while allowing task-aware corrections to refine the imputation. Moreover, it reduces the effective search space, enabling the model to focus on adjustments that improve downstream performance rather than reconstructing values from scratch.

This leads to the equivalent objective:
\begin{equation}
\min_{\theta, \boldsymbol{\delta}} 
\mathbb{E}_{(\mathbf{x}^{\text{obs}}, y)} 
\left[ \ell\big(f_\theta(\mathbf{x}^{\text{obs}}, \hat{\mathbf{x}}^{\text{u}} + \boldsymbol{\delta}), y\big) \right]
+ g(\boldsymbol{\delta}).
\end{equation}

A simple choice is a quadratic regularization:
\begin{equation}
g(\boldsymbol{\delta}) = \lambda \|\boldsymbol{\delta}\|^2,
\end{equation}
which encourages imputations to remain close to the reference.


\paragraph{Unified imputation operator.}
We define a general imputation operator:
\begin{equation}
\mathbf{x}^{\text{u}} = \mathcal{I}_\psi(\mathbf{x}^{\text{obs}}, \mathbf{m}, \hat{\mathbf{x}}^{\text{u}}),
\end{equation}
where $\psi$ parameterizes how imputations are computed.

This formulation yields a unified objective:
\begin{equation}
\min_{\theta, \psi} 
\mathbb{E}_{(\mathbf{x}^{\text{obs}}, y)} 
\left[
\ell\big(f_\theta(\mathbf{x}^{\text{obs}}, \mathcal{I}_\psi(\cdot)), y\big)
\right]
+ g(\mathcal{I}_\psi(\cdot)).
\end{equation}

Different choices of $\psi$ recover different instantiations:

\begin{itemize}
\item \textbf{Amortized imputation (Joint-AmortizedImpute):}
We parameterize the imputation operator as $\mathcal{I}_\phi$, with shared parameters $\phi$ that predict imputations from inputs. This can be interpreted as learning to approximate the optimal latent corrections $\boldsymbol{\delta}$ in a single forward pass, enabling generalization to unseen data and reducing computational cost. The model is trained by minimizing:
\begin{equation}
\mathcal{L} = \mathcal{L}_{\text{task}} + \lambda_{\text{pen}} \, \mathcal{L}_{\text{pen}} + \lambda_{\text{rec}} \, \mathcal{L}_{\text{rec}}, \label{eq:joint_loss}
\end{equation}
where $\lambda_{\text{pen}}, \lambda_{\text{rec}} \geq 0$, $\mathcal{L}_{\text{pen}}$ is a regularization term and $\mathcal{L}_{\text{rec}}$ is a reconstruction loss.

\item \textbf{Non-amortized imputation (Joint-LatentImpute):}
We directly optimize per-sample latent variables $\{\boldsymbol{\delta}_i\}_{i=1}^n$. The objective reduces to Eq.~\ref{eq:joint_loss} with $\lambda_{\text{rec}} = 0$.
\end{itemize}

Full methodological and implementation details of Joint-AmortizedImpute and Joint-LatentImpute are provided in Appendix~\ref{app_jai} and Appendix~\ref{app_jli}, respectively.

\section{Experimental Setup}

We evaluate the proposed framework under two complementary settings reflecting different practical scenarios regarding the availability of missing values at inference time. 
In our experiments, we focus on \emph{classification tasks}, therefore we use the Cross-Entropy (CE) loss. For more details on the classifier, refer to Appendix~\ref{classifier}.

\paragraph{Joint-AmortizedImpute: missingness at inference time.} This constitutes our main evaluation setting. 
We consider three scenarios to assess robustness under distribution shifts in missingness:

\begin{itemize}

\item \textbf{Robustness under train/test mismatch.}

\begin{itemize}
\item \textbf{Same mechanism, different missing rates:} The missingness mechanism is fixed (either MCAR, MAR or MNAR), whereas the missing rate differs between training $m_r^\text{train} \in \{0.1, 0.3\}$ and testing $m_r^\text{test}  \in \{0.1, 0.3, 0.5, 0.7\}: m_r^\text{test} \neq m_r^\text{train}$.

\item \textbf{Different mechanism, same missing rate:} The missing rate is fixed ($m_r^\text{train} =  m_r^\text{test}  \in \{0.1, 0.5\}$), while the missingness mechanism varies between training and testing (MCAR, MAR or MNAR).
\end{itemize}

\item \textbf{Overall stability.} Training and test data share the same missingness mechanism and rate. We evaluate six missignness mechanisms-- MCAR; MAR and its sequential variant; MNAR, and its latent-factor and polarization variants-- across five missing rates, $m_r \in\{0.1, 0.3, 0.5, 0.7, 0.8\}$.

\end{itemize}

\paragraph{Joint-LatentImpute: no missingness at inference time.} This setting isolates the effect of task-aware imputation during training while reflecting scenarios where training data is incomplete but deployment data is clean.
We evaluate across the same six missingness mechanisms (MCAR; MAR and its sequential variant; and MNAR with latent-factor and polarization variants) and five missing rates $m_r \in \{0.1, 0.3, 0.5, 0.7, 0.8\}$.

Please, refer to Appendix~\ref{missingness} for more details on data splits and simulated missingness mechanisms.

\paragraph{Baselines.}
We compare the proposed approach with the following imputation methods : mean/median for continuous variables, mode for categorical variables, $K$-NN using the mean of the nearest neighbors for numerical variables and mode imputation for categorical ones, MissForest and TabImpute.
We also consider the complete data baseline, which provides an upper bound on performance and allows us to quantify the impact of missingness and imputation.

\paragraph{Reference imputation choice.}
We select the reference imputation $\hat{\mathbf{x}}^{\text{u}}$ using a task-driven procedure: each candidate method is used to impute the training data, a downstream classifier is trained, and the method yielding the \emph{best validation performance} is chosen. This selection is performed per dataset and missingness setting and provides a strong initialization for the proposed method, improving stability and focusing learning on task-relevant corrections.

\paragraph{Evaluation metrics.} 
Imputation quality is measured using Root Mean Squared Error (RMSE) for numerical features and error rate for categorical features (Cat. Err.). To assess downstream utility, we train a classifier on the imputed data and report accuracy/F1-score. We report the performance gap relative to the complete-data baseline:
\begin{equation}
\mathrm{Gap} = \mathrm{Accuracy}_{\text{complete-data}} - \mathrm{Accuracy}_{\text{method}}.
\end{equation}
Lower values indicate better performance, with a gap of zero corresponding to perfect recovery. We additionally measure performance recovery using the \emph{recovery ratio}, defined as the ratio between the accuracy obtained under missing data and the accuracy achieved on the corresponding complete-data baseline:
\begin{equation}
\mathrm{RR} = \frac{\mathrm{Accuracy}_{\text{method}}}{\mathrm{Accuracy}_{\text{complete-data}}}.
\end{equation}
A value of $\mathrm{RR} = 1$ indicates full recovery of complete-data performance, while lower values reflect degradation due to missingness. Values greater than 1 indicate that the method outperforms the complete-data baseline. We define the hardness of a scenario based on the recovery ratio as:
\begin{equation}
\mathrm{Hardness} = 1 - \mathrm{RR}.
\end{equation} We define four hardness levels: \textbf{Easy} ($\mathrm{Hardness} < 0.02$), \textbf{Mild} ($0.02 \leq \mathrm{Hardness} < 0.05$), \textbf{Hard} ($0.05 \leq \mathrm{Hardness} < 0.1$), and \textbf{Severe} ($\mathrm{Hardness} \geq 0.1$).

\paragraph{Best classical method.}
For reporting purposes, we define the \emph{best classical method} in each scenario as the imputation approach achieving the highest test accuracy. This definition is used only for post-hoc comparison and does not affect model selection, which is performed using validation performance.

\paragraph{Datasets and data preprocessing.}
We evaluate our approach on classification datasets from OpenML \cite{JMLR:v22:19-920}.
For experiments involving distribution shifts (cross-mechanism and varying missing-rate), we use a set of 36 datasets with imbalance ratio below 10 to avoid severely skewed class distributions. In this setting, classification performance is evaluated using accuracy. 
To assess overall robustness and stability, including hardness analysis unrelated to distribution shifts, we extend the evaluation to a broader collection of 98 datasets.
All methods are evaluated on identical data splits to ensure fair comparison. Numerical features are standardized using z-score normalization, where statistics are computed using only observed values from the training set. Further details on dataset characteristics and domains are provided in Appendix~\ref{datasets}.

\paragraph{Statistical analysis.}
To assess statistical significance across multiple datasets, we follow standard protocols for comparing learning algorithms. Methods are ranked per dataset and compared using the Friedman test, followed by the Nemenyi post-hoc test to identify significant differences in average rank. We visualize results using critical difference (CD) diagrams, where lower ranks indicate better performance and horizontal bars connect methods that are not significantly different. 
Additionally, we report Spearman rank correlations to analyze the relationship between reconstruction quality and downstream performance. All statistical tests are conducted at a significance level of $\alpha = 0.05$.

\section{Results}

\subsection{Robustness under difficult scenarios}

\subsubsection{Train/test mismatch}

\begin{itemize}

\item \textbf{Different missing rates}

As shown in Figure~\ref{fig:recovery_cd_combined}, the proposed Joint-AmortizedImpute consistently achieves the highest recovery ratio (0.8181), substantially outperforming all classical baselines (0.69–0.70). 
The CD diagram across all scenarios reveals that the joint model achieves the best average rank and is separated from the classical methods by statistically significant differences. The results in Table~\ref{tab:joint_train_test_missingness} reveal a clear dependence on the magnitude of the train–test missingness shift. When the shift is small or when the test missingness is comparable to (or lower than) the training missingness, Joint-AmortizedImpute underperforms the best classical baseline. In contrast, when the test missingness is substantially higher than the training missingness, the proposed approach consistently provides strong gains: for example, in the 0.1 $\rightarrow$ 0.5 and 0.3 $\rightarrow$ 0.7 settings, it achieves positive improvements exceeding 0.10 on average and wins in over 80\% of cases.

\begin{figure}[h!]
\centering

\begin{subfigure}{0.4\linewidth}
\centering
\small
\label{tab:recovery_ratio}
\begin{tabular}{lc}
\toprule
\textbf{Method} & \textbf{Recovery Ratio} ($\uparrow$)\\
\midrule
Joint-AmortizedImpute (Ours) & \textbf{0.8181} \\
Column Median      & 0.7048 \\
K-NN               & 0.7043 \\
Column Mean        & 0.7038 \\
MissForest         & 0.6915 \\
TabImpute          & 0.6906 \\
\bottomrule
\end{tabular}

\caption{Average recovery ratio relative to the complete data baseline.}
\end{subfigure}
\hfill
\begin{subfigure}{0.45\linewidth}
\centering
\includegraphics[width=\linewidth]{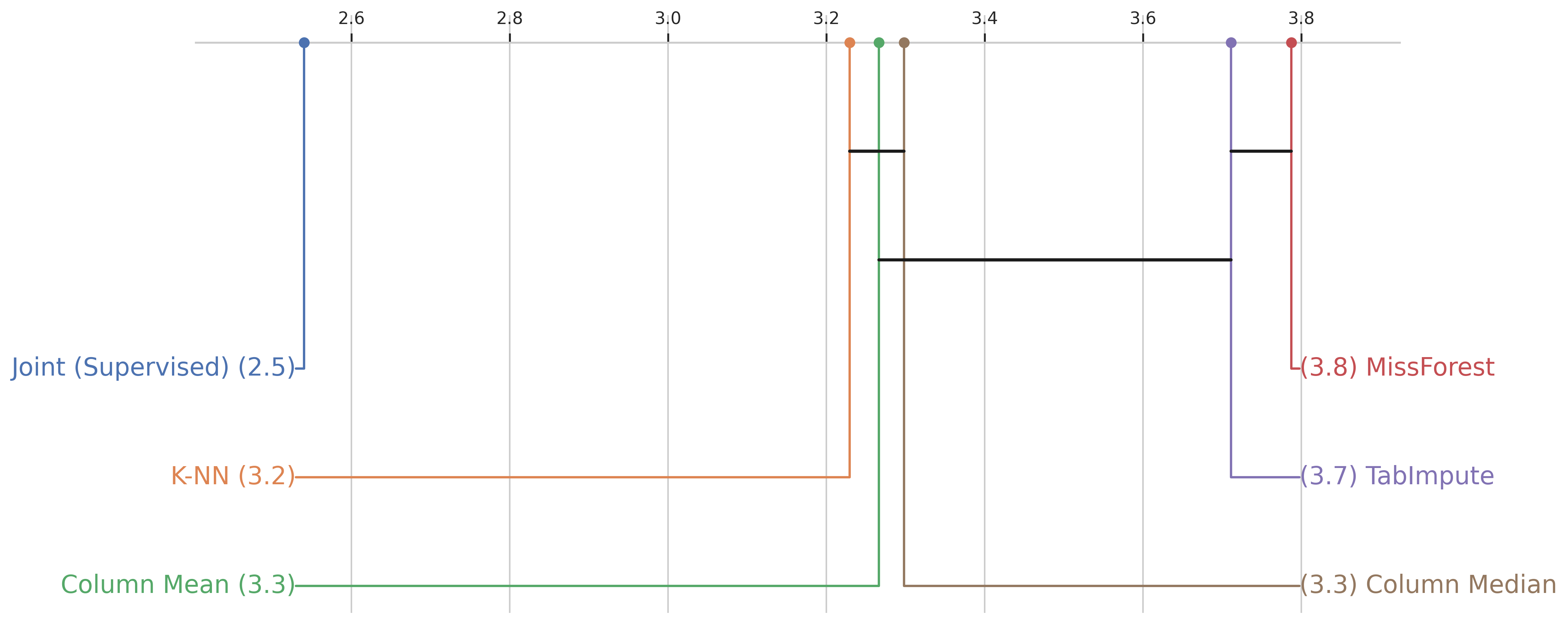}
\caption{CD diagram. }
\end{subfigure}

\caption{Comparison of imputation methods under varying missingness probabilities. The recovery ratio (left) shows that Joint-AmortizedImpute achieves the best overall performance, while the CD diagram (right) confirms statistically significant ranking differences.}
\label{fig:recovery_cd_combined}

\end{figure}

\begin{table}[h!]
\centering
\caption{Performance of Joint imputation relative to the best classical method under train--test missingness shifts. We report mean accuracy difference, win rate, and degradation rate across scenarios. }
\begin{tabular}{lccc}
\toprule \small
\textbf{Train $\rightarrow$ Test} & \textbf{Mean} ($\uparrow$) & \textbf{Win} ($\uparrow$) & \textbf{Degradation} ($\downarrow$) \\
\midrule
0.1 $\rightarrow$ 0.3 & -0.0151 & 25.00\% & 47.22\% \\
0.1 $\rightarrow$ 0.5 & \textbf{0.1240} & \textbf{74.07\%} & 21.30\% \\
0.1 $\rightarrow$ 0.7 & 0.0873 & 66.67\% & 28.21\% \\
0.3 $\rightarrow$ 0.1 & -0.0154 & 26.39\% & 43.06\% \\
0.3 $\rightarrow$ 0.5 & \textbf{0.1428} & \textbf{81.94\%} & 13.89\% \\
0.3 $\rightarrow$ 0.7 & \textbf{0.1063} & \textbf{82.05\%} & 12.82\% \\
\bottomrule
\end{tabular}\label{tab:joint_train_test_missingness}
\end{table}

\item \textbf{Different missing mechanisms}

The CD diagram (Figure~\ref{fig:cd_and_performance}, left) shows that Joint-AmortizedImpute achieves the best average rank across datasets compared to classical baselines, with statistically significant differences. This is reinforced by the recovery ratio results (Figure~\ref{fig:cd_and_performance}, right), where Joint-AmortizedImpute attains a substantially higher value (0.8381), with a clear margin over the second-best method (K-NN at 0.7220).

Moreover, Joint-AmortizedImpute consistently outperforms classical methods when evaluated under MNAR missingness, regardless of the training mechanism (Table~\ref{tab:joint_train_test_mechanism}). In contrast, when evaluated under MAR, the proposed approach exhibits negative mean differences and low win rates, indicating limited generalization under this setting. Interestingly, models trained under MNAR generalize well to MCAR, achieving high win rates and positive mean improvements.

The strong performance observed in MCAR/MAR $\rightarrow$ MNAR reflects that MNAR test settings are intrinsically more challenging, providing greater opportunity for task-aware imputation to improve performance. When training on MNAR and testing on MCAR, the problem becomes easier, as MCAR introduces only random missingness. On the other hand, MAR introduces structured missingness that depends on observed features, which can conflict with patterns learned under MNAR, where missingness depends on unobserved values, leading to a mismatch in how missingness relates to the predictive features. Similarly, training under MAR may encourage reliance on such structured missingness, which breaks under MCAR, resulting in degraded performance.

\begin{figure}[h!]
\centering

\begin{minipage}{0.48\linewidth}
    \centering
    \includegraphics[width=\linewidth]{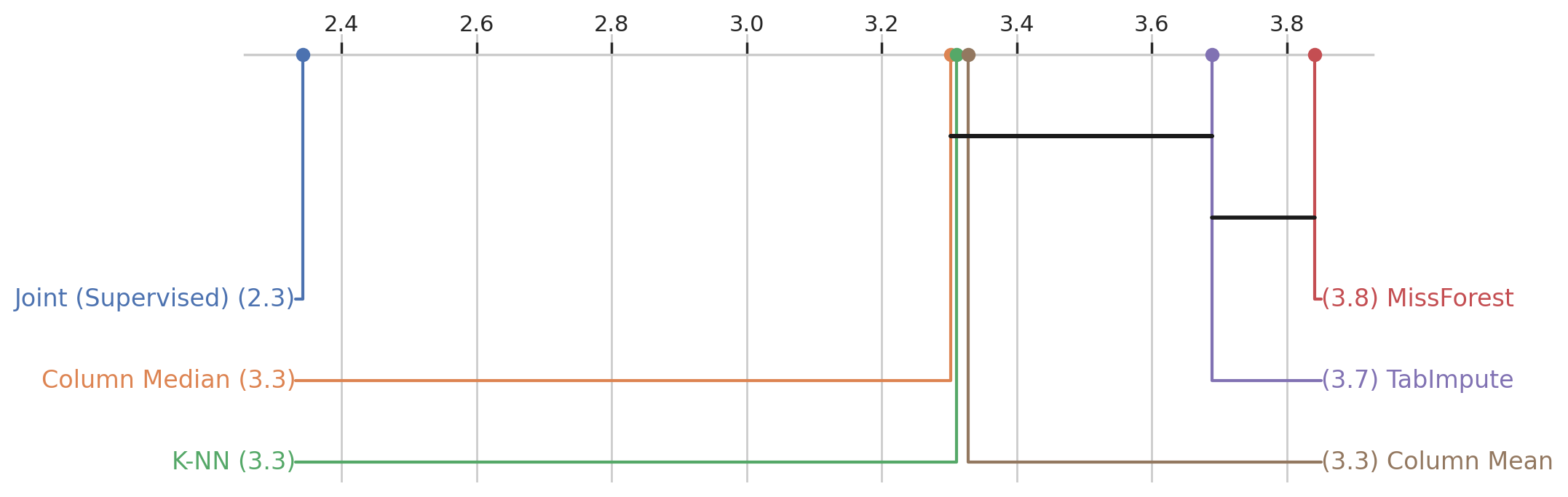}
\caption*{(a) CD diagram (Accuracy)}
\end{minipage}
\hfill
\begin{minipage}{0.48\linewidth}
    \centering
    \small
    \begin{tabular}{lcc}
    \toprule
    \textbf{Method} &  \textbf{Recovery Ratio (↑)} \\
    \midrule
    Joint-AmortizedImpute (Ours) &  \textbf{0.8381} \\
    K-NN               &  0.7220          \\
    Column Mean        & 0.7178          \\
    Column Median      &  0.7162          \\
    TabImpute          & 0.7069          \\
    MissForest         &  0.7021          \\
    \bottomrule
    \end{tabular}
    \caption*{(b) Average recovery ratio relative to the complete data baseline.}
\end{minipage}

\caption{Comparison of imputation methods.}
\label{fig:cd_and_performance}

\end{figure}

\begin{table}[h!]
\centering
\caption{Performance of Joint imputation relative to the best classical method under train--test missingness mechanism shifts. We report mean accuracy difference (Joint minus Best Classical), win rate, and degradation rate across scenarios.}
\begin{tabular}{lccc}
\toprule \small
\textbf{Train $\rightarrow$ Test} & \textbf{Mean} ($\uparrow$) & \textbf{Win} ($\uparrow$) & \textbf{Degradation} ($\downarrow$)\\
\midrule
MAR $\rightarrow$ MCAR  & -0.0249 & 23.53\% & 55.29\% \\
MAR $\rightarrow$ MNAR  & \textbf{0.1536} & \textbf{77.65\%} & 18.82\% \\
MCAR $\rightarrow$ MAR  & -0.0134 & 24.32\% & 51.35\% \\
MCAR $\rightarrow$ MNAR & \textbf{0.1783} & \textbf{79.73\%} & 13.51\% \\
MNAR $\rightarrow$ MAR  & -0.0267 & 23.46\% & 55.56\% \\
MNAR $\rightarrow$ MCAR & \textbf{0.1352} & \textbf{80.00\%} & 15.00\% \\
\bottomrule
\end{tabular}
\label{tab:joint_train_test_mechanism}
\end{table}

\end{itemize}

\subsubsection{Difficult scenarios not due to train/test mismatch}

We further analyze performance in intrinsically difficult missingness settings, defined by high hardness, i.e., scenarios where classical imputation methods fail to recover complete-data performance (see Section 5). These correspond to the “hard” and “severe” regimes in our evaluation.

\begin{figure}[h!]
    \centering
    
    \begin{subfigure}{0.48\linewidth}
        \centering
        \includegraphics[width=\linewidth]{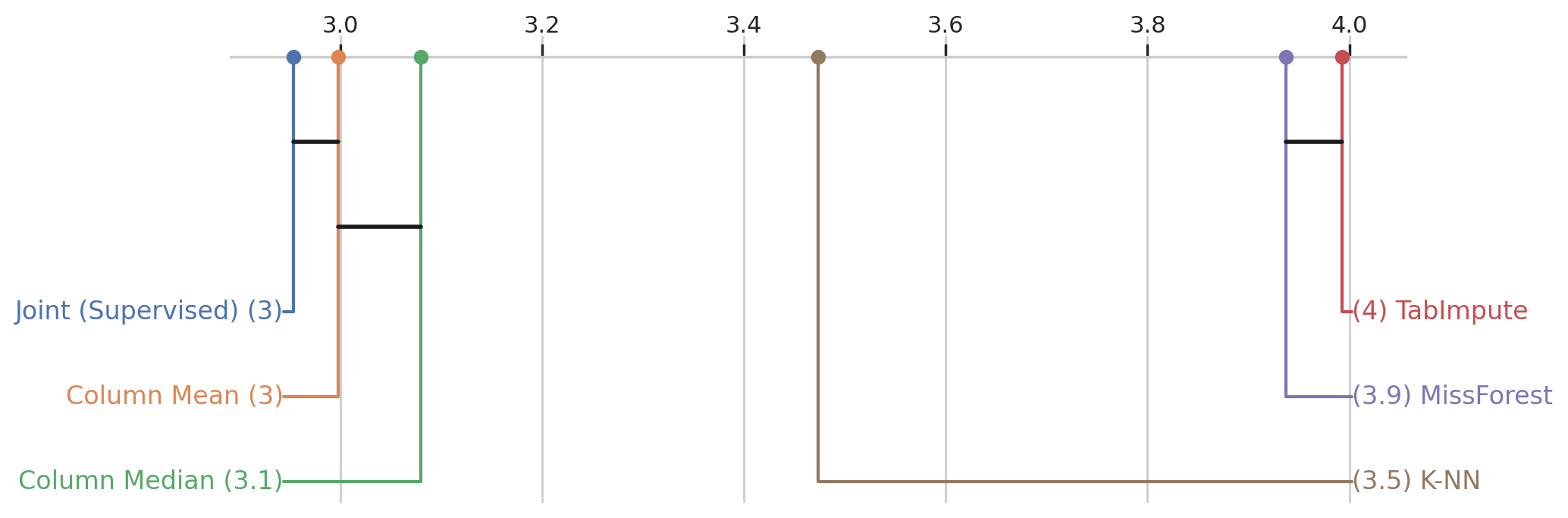}
        \caption{Accuracy}
        \label{fig:cd_acc_hard_severe_amortized}
    \end{subfigure}
    \hfill
    \begin{subfigure}{0.48\linewidth}
        \centering
        \includegraphics[width=\linewidth]{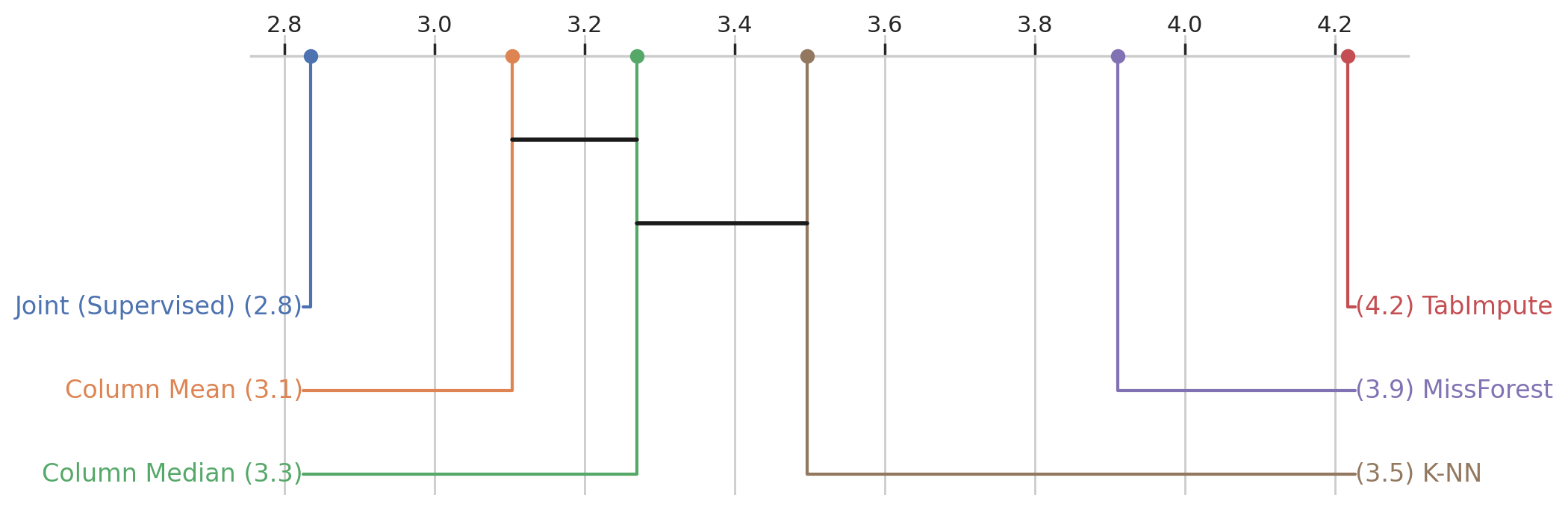}
        \caption{F1-score}
        \label{fig:cd_f1_hard_severe_amortized}
    \end{subfigure}
    
    \caption{Joint-AmortizedImpute: CD diagrams (scenario-level) for hard and severe scenarios. Lower ranks indicate better performance. }
    \label{fig:cd_hard_severe_amortized}
\end{figure}

\paragraph{Joint-AmortizedImpute} When focusing on harder missingness settings, JointAmortized-Impute achieves the lowest rank in both accuracy and F1-score. Though in terms of classification accuracy it is not significantly different from Column Mean, for F1-score it is significantly better than the remaining methods.

\begin{figure}[h!]
    \centering
    
    \begin{subfigure}[b]{0.49\linewidth}
        \centering
        \includegraphics[width=\linewidth]{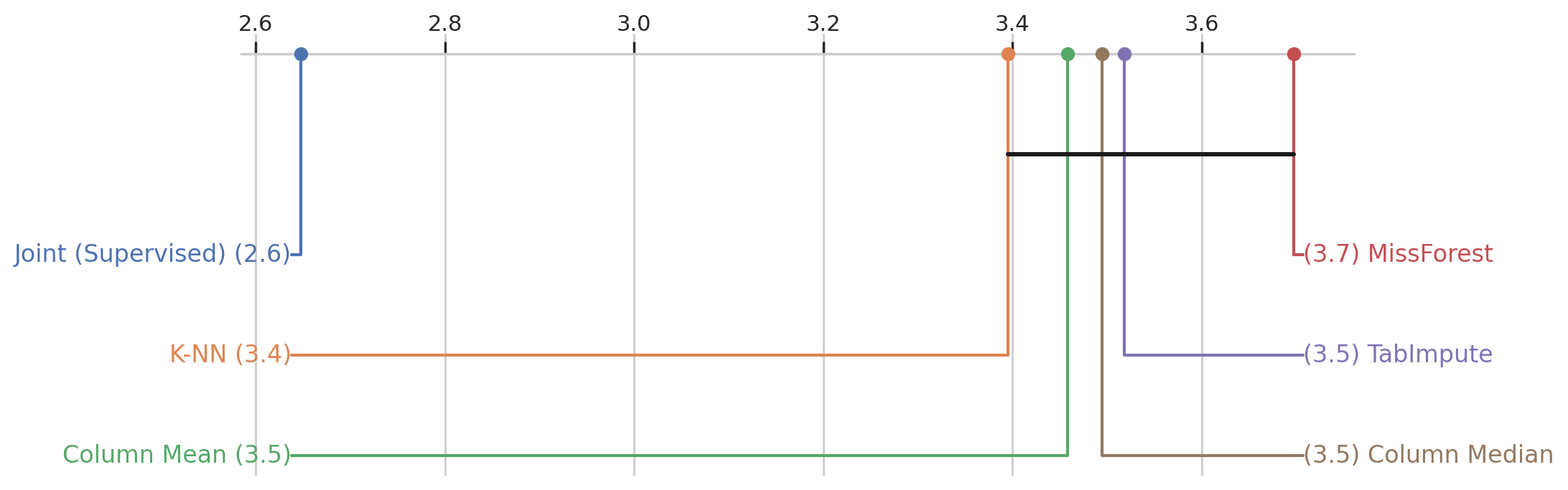}
        \caption{Accuracy}
        \label{fig:cd_acc_hard_severe}
    \end{subfigure}
    \hfill
    \begin{subfigure}[b]{0.49\linewidth}
        \centering
        \includegraphics[width=\linewidth]{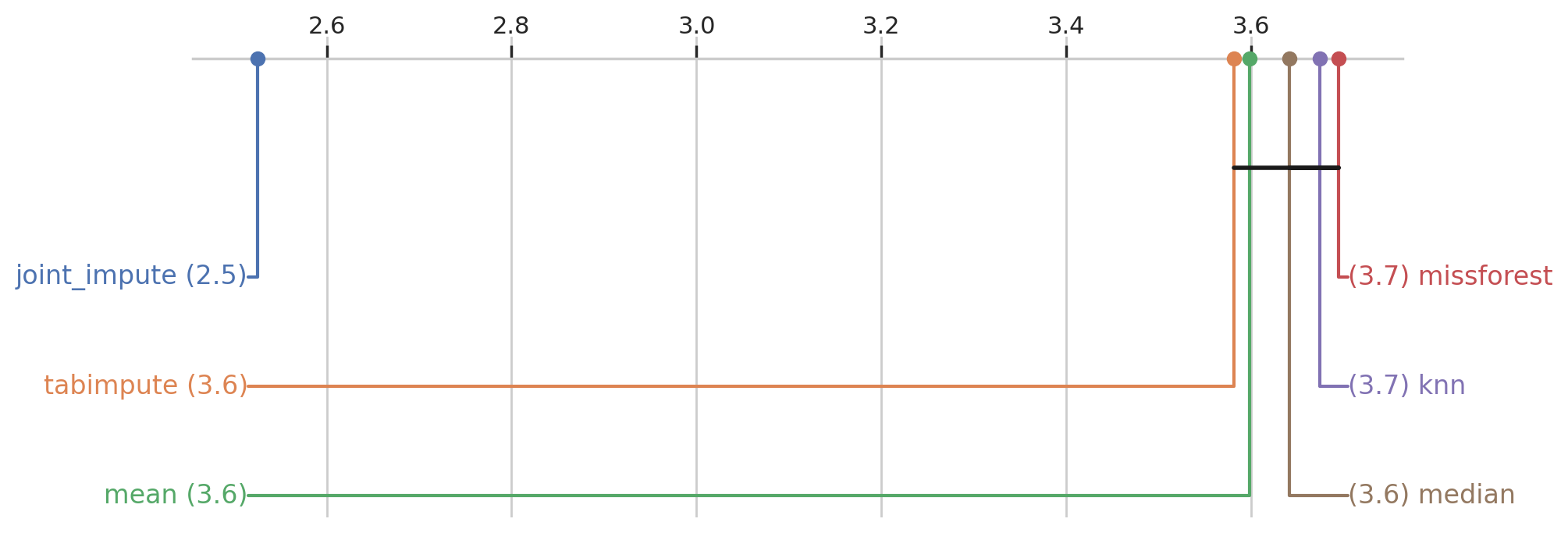}
        \caption{F1 score}
        \label{fig:cd_f1_hard_severe}
    \end{subfigure}
    \caption{Joint-LatentImpute: CD diagrams (scenario-level) for hard and severe scenarios. }
    \label{fig:cd_hard_severe}
\end{figure}

\paragraph{Joint-LatentImpute.} Figure~\ref{fig:cd_hard_severe} presents the CD diagrams restricted to hard and severe scenarios, where the proposed Joint-LatentImpute is the top-performing method across both accuracy and F1 score (statistically significant, Nemenyi post-hoc, all $p < 10^{-6}$).
Moreover, we observe a weak but statistically significant negative correlation between improvement and the performance of the selected reference imputation (Spearman, $\rho = -0.07$, $p < 0.005$).

\subsection{Reconstruction error \& its relationship to downstream performance}

Table \ref{tab:combined_imputation_ranking} reports the imputation quality ranking for both Joint-LatentImpute and Joint-AmortizedImpute across numerical (RMSE) and categorical features.

\begin{table}[h!]
\centering
\small
\begin{tabular}{lcccc}
\toprule
& \multicolumn{2}{c}{\textbf{Joint-LatentImpute}} & \multicolumn{2}{c}{\textbf{Joint-AmortizedImpute}} \\
\cmidrule(lr){2-3} \cmidrule(lr){4-5}
\textbf{Method} & \textbf{RMSE Rank }($\downarrow$) & \textbf{Cat. Err. Rank} ($\downarrow$)& \textbf{RMSE Rank} ($\downarrow$)& \textbf{Cat. Err. Rank}($\downarrow$) \\
\midrule
Joint (Ours)       & \textbf{2.96} & 1.8 & 2.80 & 2.01 \\
Column Mean        & 3.11          & \textbf{1.6} & \textbf{2.62} & \textbf{1.77} \\
K-NN               & 3.39          & \textbf{1.6} & 3.15 & -- \\
Column Median      & 4.64          & \textbf{1.6} & 3.85 & -- \\
MissForest         & 3.02          & 2.0 & 4.82 & 3.12 \\
TabImpute          & 3.87          & 3.6 & 3.74 & 3.65 \\
\bottomrule
\end{tabular}
\caption{Imputation quality ranking (lower is better).}
\label{tab:combined_imputation_ranking}
\end{table}

\paragraph{Joint-LatentImpute.}
We analyze the relationship between imputation quality and downstream classification performance using Spearman rank correlation between RMSE ranks and accuracy ranks across all scenarios. We observe a statistically significant positive correlation (Spearman, $\rho = 0.16$, $p < 0.001$), which indicates that methods achieving better reconstruction quality tend to also perform better in downstream prediction. However, this correlation is weak, suggesting that reconstruction error alone does not fully explain predictive performance. For example, MissForest achieves competitive performance in terms of numerical reconstruction error, ranking second overall. However, this advantage does not translate to downstream classification performance, where simpler methods such as mean and K-NN imputation often perform better.

\paragraph{Joint-AmortizedImpute.}
We observe a statistically significant positive correlation between RMSE ranks and accuracy ranks across all scenarios (Spearman, $\rho = 0.32$, $p < 0.001$). This reinforces that, though methods achieving better reconstruction quality tend to perform better in downstream prediction, 
reconstruction quality alone does not fully explain predictive performance.

\subsection{Overall stability assessment}

\begin{figure}[h!]
    \centering
        \begin{subfigure}{0.48\linewidth}
        \centering
        \includegraphics[width=\linewidth]{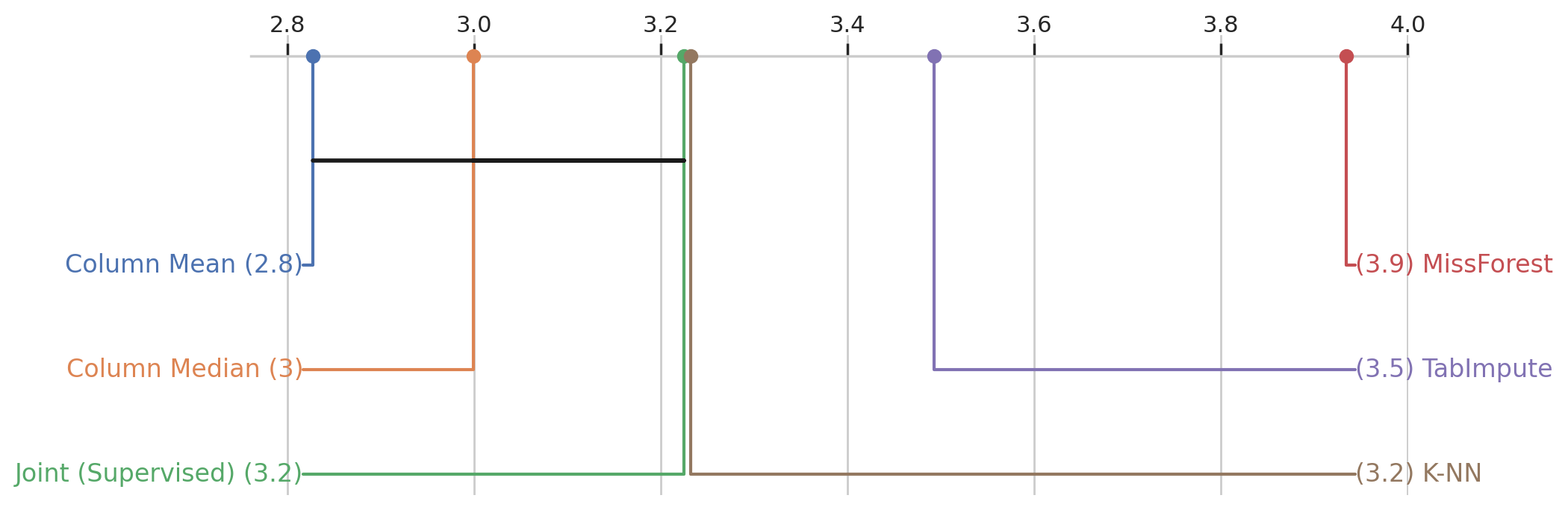}
        \caption{Accuracy.}
        \label{fig:crit}
    \end{subfigure}
    \hfill
    \begin{subfigure}{0.48\linewidth}
        \centering
        \includegraphics[width=\linewidth]{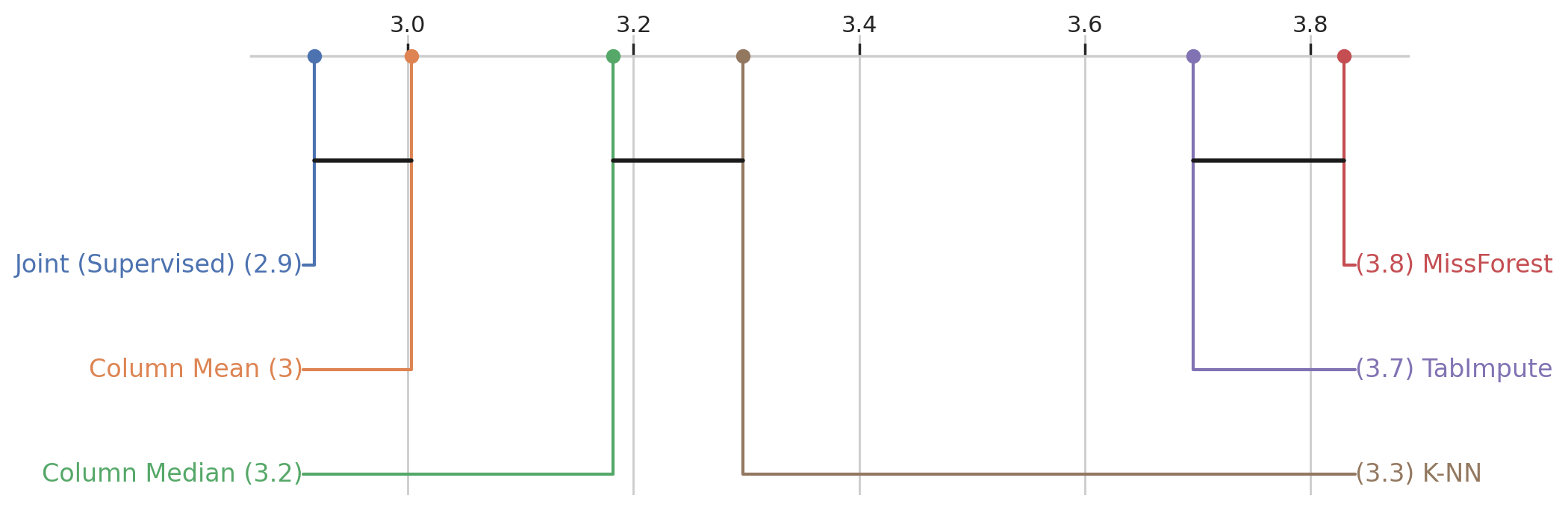}
        \caption{F1-score.}
        \label{fig:cd_f1_}
    \end{subfigure}

    \caption{Overall: CD diagrams comparing imputation methods based on performance rankings.  }
    \label{fig:cd_f1_combined}
\end{figure}

\paragraph{Joint-AmortizedImpute.}
Table~\ref{tab:recovery_ratioooo} shows that the proposed method achieves the highest average recovery ratio with respect to the complete-data baseline, highlighting its robustness in mitigating the impact of missing data compared to the benchmark imputation methods. However, on average, Joint-AmortizedImpute leads to only marginal changes relative to the selected reference imputation method, with small performance degradations (Table~\ref{tab:joint_vs_init_transposed}). 
The CD diagrams in Figures~\ref{fig:cd_f1_} and \ref{fig:crit} show that, for both accuracy and F1-score, the proposed approach is not significantly different from the best ranking approaches, Column Mean and Column Median (Nemenyi post-hoc, $p=0.38$ and $p=0.96$ for accuracy; $p=0.11$ for Column Mean in F1). 

\begin{minipage}{0.45\textwidth}
\centering
\small
\begin{tabular}{lc}
\toprule
Method & Recovery Ratio (↑) \\
\midrule
Joint-AmortizedImpute (Ours) & \textbf{0.8932} \\
Column Mean        & 0.8902 \\
K-NN               & 0.8875 \\
Column Median      & 0.8863 \\
TabImpute          & 0.8785 \\
MissForest         & 0.8593 \\
\bottomrule
\end{tabular}

\captionof{table}{Mean recovery ratio per method (higher is better).}
\label{tab:recovery_ratioooo}
\end{minipage}
\hfill
\begin{minipage}{0.45\textwidth}\small
\centering
\begin{tabular}{lcc}
\toprule
 & Accuracy (↑) & F1 (↑) \\
\midrule
Mean $\Delta$        & -0.0039 & -0.0022 \\
Median $\Delta$      & -0.0000 & -0.0003 \\
Win Rate             & 41.5\%  & 41.9\% \\
Degradation $>5\%$   & 3.8\%   & 6.0\%  \\
Degradation $>10\%$  & 0.85\%  & 1.30\% \\
\bottomrule
\end{tabular}

\captionof{table}{Joint-AmortizedImpute vs Reference performance. Positive $\Delta$ indicates improvement over the selected reference imputation method.}
\label{tab:joint_vs_init_transposed}
\end{minipage}

\paragraph{Joint-LatentImpute.}

Across all scenarios, Table~\ref{tab:recovery_ratioh} shows that the proposed joint imputation achieves the highest average recovery performance when averaged across all scenarios. Moreover, Joint-LatentImpute improves performance compared to the selected reference imputation in approximately 45\% of the cases, while it degrades performance by more than 5\% and 10\% in, respectively, 2\% and 0.3\% of the scenarios, suggesting that the proposed method is generally stable and rarely introduces large performance drops (Table~\ref{tab:perfdiff}).

\begin{table}[h!]
\centering

\begin{minipage}{0.45\linewidth}
\centering 
\caption{Joint-LatentImpute (Ours): recovery of complete data performance.} \label{tab:recovery_ratioh}
\begin{tabular}{lc}
\toprule
Method & Recovery Ratio (↑)\\
\midrule
Joint-LatentImpute (Ours) & \textbf{0.9641} \\
K-NN         & 0.9595 \\
Mean         & 0.9580 \\
TabImpute    & 0.9579 \\
MissForest   & 0.9577 \\
Median       & 0.9570 \\
\bottomrule
\end{tabular}
\end{minipage}
\hfill
\begin{minipage}{0.4\linewidth}
\centering
\small
\caption{Joint-LatentImpute (Ours) performance difference over the selected reference (Accuracy / F1).}\label{tab:perfdiff}
\begin{tabular}{lcc}
\toprule
Metric & Acc (↑) & F1 (↑)\\
\midrule
Mean $\Delta$      & -0.0007 & -0.0022 \\
Win rate           & 0.4548  & 0.4051 \\
Degradation $>5\%$         & 0.0189  & 0.0522 \\
Degradation $>10\%$        & 0.0025  & 0.0124 \\
\bottomrule
\end{tabular}
\end{minipage}
\end{table}

\section{Conclusions, Limitations and Future Work}

\paragraph{Conclusions.}
This work studies the role of task-aware imputation in supervised learning with missing data, with a focus on understanding when it improves predictive performance. 
Our results show that task-aware imputation does not uniformly improve predictive performance, but instead \emph{provides benefits primarily in challenging regimes}, particularly under distribution shift and in MNAR settings, where classical methods fail to recover complete-data performance. 
We further find that reconstruction quality is only weakly correlated with downstream performance (Spearman $\rho\approx$0.16–0.32), suggesting that optimizing imputation for reconstruction alone is insufficient. These results indicate that \emph{imputation should be treated as a task-dependent problem}, rather than a preprocessing step focused solely on data fidelity.

\paragraph{Limitations and Future Work.} 
The proposed Joint-AmortizedImpute approach uses fixed hyperparameters across all datasets. While this ensures consistent evaluation, it may limit adaptability; dataset-specific tuning or automated optimization could improve performance and robustness. 
The experimental comparison is also limited to a small set of baseline methods, and extending the benchmark would enable a more comprehensive evaluation. 
Furthermore, experiments focus exclusively on tabular classification, leaving generalization to other settings (e.g., regression or structured data such as time series) as an open question. 
Finally, the method does not explicitly model the missingness mechanism, which may reduce effectiveness when missingness is structured or informative. Incorporating mechanism-aware components is a promising direction for future work.

\paragraph{Broader impacts.} Improved handling of missing data can enhance the reliability of machine learning systems in real-world applications such as healthcare, finance, and scientific analysis. By aligning imputation with predictive objectives, the proposed approach may support better decision-making. However, task-aware imputation may also amplify biases in the observed data, leading to unfair or misleading outcomes, particularly in sensitive domains. Future work should explore fairness-aware imputation and develop safeguards to mitigate these risks.

\bibliographystyle{unsrt}
\bibliography{imputation}

\appendix

\section{Experimental setup}

All experiments were conducted using Python 3.10.19. The implementation relies on PyTorch \cite{ansel2024pytorch2}, scikit-learn, NumPy and pandas. 
Tabular foundation models are incorporated through the \texttt{tabimpute} library (downloaded from its public GitHub repository). Evaluation metrics such as accuracy and F1-score are computed using \texttt{scikit-learn}.

Experiments were conducted on a machine equipped with an NVIDIA GPU with CUDA support.

\subsection{Datasets and data preprocessing}

We restrict our evaluation to OpenML classification datasets without missing values. This allows us to generate missingness in a fully controlled manner, ensuring that the ground-truth complete data is always available.

\begin{table}[h]
\centering

\begin{minipage}{0.45\linewidth}
\centering
\begin{tabular}{lc}
\toprule
Property & Range \\
\midrule
\#Datasets & 98 \\
Samples & 350 -- 7000 \\
Features & 4 -- 49 \\
\# Categorical features & 0 -- 23 \\
Classes & 2 -- 40 \\
\bottomrule
\end{tabular}
\caption{Dataset characteristics.}\label{tab:dataset_summary_charact}
\end{minipage}
\hfill
\begin{minipage}{0.45\linewidth}
\centering
\begin{tabular}{lc}
\toprule
Domain & \#Datasets \\
\midrule
Synthetic & 19 \\
Finance & 15 \\
Medical / Healthcare & 14 \\
Computer / IT / Security & 14 \\
Social / Behavioral & 11 \\
Physical / Environmental & 11 \\
Biology / Chemistry & 8 \\
Industrial / Engineering & 6 \\
\bottomrule
\end{tabular}
\caption{Dataset domains (manual annotation).}\label{tab:dataset_summary_domain}
\end{minipage}

\end{table}

Tables~\ref{tab:dataset_summary_charact} and \ref{tab:dataset_summary_domain} summarize, respectively the domains and characteristics of the 98 datasets used in our main evaluation. The collection spans a wide range of domains, dataset sizes, feature dimensionalities, and class cardinalities, ensuring a diverse and representative benchmark. In particular, datasets include both low- and high-dimensional settings, varying proportions of categorical features, and both binary and multi-class classification tasks.

\textbf{List of datasets:}

\begin{itemize}
\item 1StudentPerfromance
\item Accidents\_Prediction\_Dataset
\item Australian\_seed\_1\_nrows\_2000\_nclasses\_10\_ncols\_100\_stratify\_True
\item Bank-Note-Authentication-UCI
\item Bank\_Customer\_Churn
\item CanadaFiresD
\item Creditability-German-Credit-Data
\item Heart-Disease-Dataset-(Comprehensive)
\item Intersectional-Bias-Assessment-(Training-Data)
\item JapaneseVowels
\item LED-display-domain-7digit
\item LEV
\item Mobile\_Price
\item Multiclass\_Classification\_for\_Corporate\_Credit\_Ratings
\item Phishing\_Legitimate\_full
\item PieChart2
\item PizzaCutter1
\item Pumpkin\_Seeds
\item SWD
\item Satellite
\item Titanic
\item VulNoneVul
\item ada
\item air-quality-and-pollution-assessment
\item analcatdata\_supreme
\item arsenic-female-bladder
\item aztrees3
\item balance-scale
\item bank-marketing\_seed\_1\_nrows\_2000\_nclasses\_10\_ncols\_100\_stratify\_True
\item bank8FM
\item blood-transfusion-service-center
\item boston
\item bwin\_amlb
\item cardiotocography
\item churn
\item climate-model-simulation-crashes
\item collins
\item compas-two-years
\item covertype\_seed\_0\_nrows\_2000\_nclasses\_10\_ncols\_100\_stratify\_True
\item cpu\_small
\item credit-approval
\item credit-g
\item delta\_ailerons
\item diabetes
\item dis
\item doa\_bwin
\item electrical\_grid\_stability\_simulated\_classification
\item eucalyptus
\item eye\_movements
\item glioma\_grading\_clinical\_and\_mutation\_features
\item hazelnut-spread-contaminant-detection
\item heloc
\item hepatitis\_c\_virus\_hcv\_for\_egyptian\_patients
\item house\_16H\_seed\_0\_nrows\_2000\_nclasses\_10\_ncols\_100\_stratify\_True
\item ibm-employee-performance
\item ilpd
\item jungle\_chess\_2pcs\_raw\_endgame\_complete\_seed\_4\_nrows\_2000\_nclasses\_10\_ncols\_100\_stratify\_True
\item kc2
\item kdd\_ipums\_la\_97-small
\item mabbob\_ela\_as\_5d\_classify
\item maternal\_health\_risk
\item mfeat-zernike
\item numerai28.6\_seed\_0\_nrows\_2000\_nclasses\_10\_ncols\_100\_stratify\_True
\item oil\_spill
\item page-blocks
\item phoneme
\item pollen
\item puma8NH
\item qsar-biodeg
\item red\_wine
\item ringnorm
\item rl
\item rmftsa\_ladata
\item road-safety\_seed\_0\_nrows\_2000\_nclasses\_10\_ncols\_100\_stratify\_True
\item segment
\item seismic-bumps
\item shrutime
\item sick
\item socmob
\item steel-plates-fault
\item stock
\item strikes
\item student\_lifestyle\_dataset
\item sylvine\_seed\_1\_nrows\_2000\_nclasses\_10\_ncols\_100\_stratify\_True
\item texture
\item thyroid-allbp
\item thyroid-allhyper
\item tokyo1
\item twonorm
\item vehicle
\item visualizing\_soil
\item vowel
\item waveform
\item wdbc
\item wilt
\item wind
\item wine-quality-red
\item yeast
\end{itemize}

\begin{table}[h!]
\centering

\begin{minipage}{0.48\textwidth}
\centering
\begin{tabular}{l c}
\toprule
\textbf{Metric} & \textbf{Range} \\
\midrule
\# Datasets & 36 \\
Samples & 350--7000 \\
Features & 4--48 \\
\# Categorical features & 0--15 \\
Classes & 2--10 \\
\bottomrule
\end{tabular}
\caption{Dataset characteristics.}\label{tab:dataset_summary_char_subset}
\end{minipage}
\hfill
\begin{minipage}{0.48\textwidth}
\centering
\begin{tabular}{l r}
\toprule
\textbf{Domain} & \textbf{\# Datasets} \\
\midrule
Synthetic & 6 \\
Finance & 11 \\
Medical / Healthcare & 5 \\
Computer / IT / Security & 4 \\
Social / Behavioral & 4 \\
Physical / Environmental & 4 \\
Biology / Chemistry & 1 \\
Industrial / Engineering & 1 \\
\bottomrule
\end{tabular}\caption{Dataset domains (manual annotation).}\label{tab:dataset_summary_domain_subset}
\end{minipage}
\end{table}

\label{datasets}

For experiments involving distribution shifts, we consider a set of 36 datasets, whose characteristics and domains are reported in Tables~\ref{tab:dataset_summary_char_subset} and \ref{tab:dataset_summary_domain_subset} . These datasets are selected to ensure stable evaluation under more challenging settings, such as cross-mechanism and varying missing-rate scenarios, while maintaining diversity across domains.

\textbf{List of datasets:}

\subsection{Datasets with reported win-rate results}

\begin{itemize}
\item delta\_ailerons
\item compass\_seed\_1\_nrows\_2000\_nclasses\_10\_ncols\_100\_stratify\_True
\item electrical\_grid\_stability\_simulated\_classification
\item cmc
\item Mobile\_Price
\item covertype\_seed\_0\_nrows\_2000\_nclasses\_10\_ncols\_100\_stratify\_True
\item cpu\_small
\item diabetes\_risk\_prediction\_dataset
\item compas-two-years
\item balance-scale
\item LED-display-domain-7digit
\item Bank\_Customer\_Churn
\item Multiclass\_Classification\_for\_Corporate\_Credit\_Ratings
\item JapaneseVowels
\item Pumpkin\_Seeds
\item Titanic
\item analcatdata\_supreme
\item Phishing\_Legitimate\_full
\item 1StudentPerfromance
\item Australian\_seed\_1\_nrows\_2000\_nclasses\_10\_ncols\_100\_stratify\_True
\item Bank-Note-Authentication-UCI
\item credit
\item diabetes
\item credit-approval
\item Heart-Disease-Dataset-(Comprehensive)
\item Intersectional-Bias-Assessment-(Training-Data)
\item KDDCup09\_upselling
\item bank-marketing\_seed\_1\_nrows\_2000\_nclasses\_10\_ncols\_100\_stratify\_True
\item ada
\item bank8FM
\item CanadaFiresD
\item VulNoneVul
\item blood-transfusion-service-center
\item boston\_corrected
\item credit-g
\item bwin\_amlb
\end{itemize}

\subsection{Data splits and missingness}

Each dataset is split into training (70\%), validation (15\%), and test (15\%) sets using stratified sampling with respect to the class labels. Splits are generated using a fixed random seed (42) to ensure reproducibility.
Categorical variables are encoded as integer category codes prior to missingness generation.
Missing values are introduced after splitting the data, independently for the training, validation and test sets.

Other than the classical MCAR, MAR, and MNAR settings, we consider three additional missingness mechanisms capturing different dependency structures. These mechanisms are implemented using the TabImpute framework \cite{feitelberg2025tabimputeaccuratefastzeroshot}, which provides configurable generators for structured missingness patterns.
\label{missingness}

MAR Sequential is generated using an $\epsilon$-greedy bandit process, where missingness depends on previously observed values. MNAR latent factor missingness is driven by a low-rank latent representation of the data, making missingness depend on unobserved variables. In particular, the latent rank is randomly sampled between 1 and 5, introducing variability in the underlying dependency structure. MNAR Polarization missingness preferentially removes values in the lower tail of the distribution (below the 25th percentile), simulating censoring effects.

\subsection{Benchmark imputation methods}

For each dataset and missingness configuration, we load both the fully observed data and the corresponding corrupted version with missing values. A binary mask indicating observed entries is also provided.
Numerical features are standardized using a StandardScaler fitted on the training data and applied to all splits.

We evaluate five imputation methods: mean, median, $K$-NN, MissForest and TabImpute. Numerical and categorical features are handled separately when required, with categorical variables typically imputed using the mode. If a categorical feature contains only missing values, a default value (0) is assigned.

Numerical features are standardized using a \texttt{StandardScaler} fitted on the training data with missing values. The same transformation is applied to validation and test sets.

Categorical variables are represented as integer-encoded categories during imputation. After imputation, all features are transformed using one-hot encoding before being passed to the classifier.

Each imputation method is applied independently to the training, validation, and test sets. For methods requiring fitting, the imputer is trained on the training data and then applied to validation and test sets.

Specifically:
\begin{itemize}
    \item \textbf{Mean and median imputation} are computed exclusively on the training data. For numerical features, the mean or median of each feature is calculated using the training set and used to fill missing values. For categorical features, the mode is used. The same statistics are then consistently applied to validation and test sets through feature encoding alignment.

    \item \textbf{$K$-NN imputation} (implemented with \texttt{sklearn.impute.KNNImputer}) is fitted on the numerical features of the training data only. Missing numerical values are imputed based on nearest neighbors in the training set with $K=5$. Categorical features are imputed separately using the training-set mode.

    \item \textbf{MissForest} (implemented via the \texttt{missforest} package \url{https://github.com/yuenshingyan/MissForest}) is trained exclusively on the training data. A random forest model is iteratively fitted to predict missing values using observed features, and the trained imputer is then used to reconstruct the training data.

    \item \textbf{TabImpute} (using \texttt{tabimpute.interface.TabImputeCategorical} \cite{feitelberg2025tabimputeaccuratefastzeroshot}) is applied independently to each data split. The same hyperparameters are used across splits, but the model does not reuse information from the training set when imputing validation or test data. TabImpute processes data in chunks, with a maximum of 64 rows per chunk and up to 4 chunks per batch. 
\end{itemize}

After imputation, categorical variables are one-hot encoded, and a classifier is trained on the imputed training data.

For each dataset and missingness configuration, the imputation method yielding the highest validation accuracy is selected.
Imputation quality is assessed by comparing imputed values to ground truth using a masked error metric computed only on missing entries.
Imputation time is recorded to assess computational efficiency.
For reproducibility, results and best-performing imputations are stored for each dataset, mechanism, and missingness level.

\subsection{Classifier} For all experiments, we use a simple feedforward neural network to isolate the effect of the proposed imputation method. The classifier consists of a single hidden layer with ReLU activation:
\begin{equation}
f_\theta(x) = W_2 \, \sigma(W_1 x + b_1) + b_2,
\end{equation}
where $\sigma(\cdot)$ denotes the ReLU activation.

\label{classifier}
The hidden dimension is fixed to $h=32$ across all experiments. The classifier is intentionally kept lightweight to ensure that performance differences are primarily attributable to the imputation strategy rather than model capacity.

\subsection{ Joint-LatentImpute}

\label{app_jli}

For numerical features, additive offsets are learned for missing entries. For categorical features, missing values are represented as learnable logits over category distributions. The model is initialized with reference imputations obtained from the best-performing baseline method (selected via validation). These reference values serve as anchors, and the model learns deviations from them. 

Let $\mathcal{M}$ denote the set of missing entries. A reference imputation $Z^{ref}$ is computed beforehand using the best-performing baseline method, and is used to initialize missing values.

Let $x_{ij}$ denote a missing numerical entry. It is parameterized as:
\begin{equation}
x_{ij} = z^{ref}_{ij} + \alpha \cdot \delta_{ij},
\end{equation}
where $\delta_{ij}$ is a learnable parameter and $\alpha = 0.1$.
Observed entries remain unchanged throughout training.

Each categorical variable is represented using a smoothed one-hot encoding. For missing entries, logits are defined as:
\begin{equation}
\ell_{ij} = \log p^{ref}_{ij} + \alpha \cdot \delta_{ij},
\end{equation}
where $p^{ref}_{ij}$ is the reference categorical distribution.

For missing entries, perturbations $\delta_{ij}$ are initialized with small Gaussian noise to avoid degenerate initialization and encourage exploration of the imputation space.

Sampling is performed using the Gumbel-Softmax relaxation:
\begin{equation}
\tilde{x}_{ij} = \text{GumbelSoftmax}(\ell_{ij}, \tau),
\end{equation}
which enables differentiable sampling of discrete variables.

The model is trained by minimizing:
\begin{equation}
\mathcal{L} = \mathcal{L}_{CE} + \lambda_{pen} \mathcal{L}_{pen} = \mathcal{L}_{CE} + \lambda_{num} \mathcal{L}_{num} + \lambda_{cat} \mathcal{L}_{cat},
\end{equation}
where $\lambda_{num} = 0.1$ and $\lambda_{cat} = 0.5$.

\begin{itemize}
    \item Classification Loss\begin{equation}
\mathcal{L}_{CE} = - \sum_{i} \log p(y_i \mid x_i)
\end{equation}
\item Numerical Regularization\begin{equation}
\mathcal{L}_{num} = \frac{1}{|\mathcal{M}_{num}|} \sum_{(i,j)\in \mathcal{M}_{num}} \delta_{ij}^2
\end{equation}
\item Categorical Regularization
\begin{equation}
\mathcal{L}_{cat} = \frac{1}{|\mathcal{M}_{cat}|} \sum_{(i,j)\in \mathcal{M}_{cat}} 
\mathrm{KL}\big(p_{\delta_{ij}} \,\|\, p^{ref}_{ij}\big)
\end{equation}

\end{itemize}

These regularization terms constrain imputations to remain close to the reference solution.

Training is performed using mini-batch stochastic gradient descent with the Adam optimizer. Two optimizers are used:
\begin{itemize}
    \item one for model parameters $\theta$,
    \item one for latent variables $\Delta$, with a learning rate reduced by a factor of 10.
\end{itemize}

A temperature annealing schedule is applied:
\begin{equation}
\tau_t = \max(\tau_{\min}, \tau_0 \cdot 0.995^t),
\end{equation}
with $\tau_0 = 1.0$ and $\tau_{\min} = 0.2$.

Training proceeds for a maximum of 1000 epochs with batch size 64.

To improve robustness, the following mechanisms are used:
\begin{itemize}
    \item Early stopping based on validation accuracy (patience = 50);
    \item Gradient clipping with maximum norm 5.
\end{itemize}

After training, missing values are reconstructed as follows:

\begin{itemize}
    \item Numerical
    \begin{equation}
x_{ij} = z^{ref}_{ij} + \alpha \cdot \delta_{ij}
\end{equation}
\item Categorical
\begin{equation}
\hat{c}_{ij} = \arg\max p_{\delta_{ij}}
\end{equation}
\end{itemize}

\paragraph{Implementation details.} Model parameters are initialized using Kaiming uniform initialization for linear layers. Missing numerical entries are initialized from the reference values with additive Gaussian noise $\mathcal{N}(0, 0.1^2)$. For categorical features, missing values are initialized by sampling from a noisy version of the reference distribution, obtained via a softmax over perturbed log-probabilities. Observed categorical values are encoded as smoothed one-hot vectors.
Optimization is performed using Adam with separate learning rates for model parameters and latent variables: respectively, $1\times10^{-3}$ and $1\times10^{-4}$. Gradients are clipped to a maximum norm of 5 to ensure stability. Early stopping is applied with a patience of 50 iterations.
Mini-batch training is used with batch size 64. The Gumbel-Softmax temperature is annealed exponentially during training from 1.0 to a minimum of 0.2.
All experiments are conducted using a fixed random seed (42) for reproducibility.

\paragraph{Number of parameters.}
Let $P_{\text{classifier}}$ be the number of parameters of the downstream classification model, $d_{\text{num}}$ the number of missing numerical entries, and $d_{\text{cat}}$ the number of latent parameters associated with categorical missing entries. 
The total number of parameters is given by
\begin{equation}
\# \text{Parameters} = P_{\text{classifier}} + d_{\text{num}} + d_{\text{cat}}.
\end{equation}

\paragraph{Computational Complexity.}
Let $n$ be the number of samples, $d$ the input dimensionality, $h$ the hidden dimension of the classifier, and $T$ the number of training iterations. 
In this approach, missing values are treated as free optimization variables, resulting in $\mathcal{O}(|\mathcal{M}|)$ additional parameters, where $\mathcal{M}$ denotes the set of missing entries.
Each training iteration requires a forward and backward pass through the classifier, with complexity $\mathcal{O}(dh + h^2)$ per sample, as well as updates to the latent variables, which incur an additional cost proportional to $|\mathcal{M}|$.

The overall training complexity is therefore:
\begin{equation}
\mathcal{O}\left(T \cdot \left(\frac{n}{b} \cdot (dh + h^2) + |\mathcal{M}|\right)\right),
\end{equation}
which becomes $\mathcal{O}(T \cdot nd)$ in the worst case.
The space complexity is $\mathcal{O}(nd + |\mathcal{M}|)$, dominated by storing the dataset and the latent variables.

\subsection{Joint-AmortizedImpute}

\label{app_jai}
Given an input with missing values $X$, a mask $M \in \{0,1\}^{d}$ indicating observed entries, and a reference imputation $Z_{\text{ref}}$, the imputation network produces a completed representation:
\[
\hat{X} = M \odot X + (1 - M) \odot \left(Z_{\text{ref}} + \alpha \cdot \Delta(X, M, Z_{\text{ref}})\right),
\]
where $\Delta(\cdot)$ is a learned residual correction and $\alpha$ is a fixed scaling factor. This formulation can be interpreted as learning task-aware deviations from a strong reference imputation, rather than predicting missing values from scratch.

Reference values $Z_{\text{ref}}$ are obtained from the best-performing baseline imputation method (selected via validation), and serve as strong initial estimates that the model refines.

Numerical features are used directly after standardization. Categorical variables are encoded as one-hot vectors. Missing categorical values are initialized using reference distributions with added noise and refined during training using a Gumbel-Softmax relaxation to enable differentiable sampling.

Cross-entropy loss is applied to the classifier predictions. For naturally missing values, the imputed outputs are regularized toward the reference imputations:
\begin{itemize}
    \item Mean squared error for numerical features
    \item KL divergence for categorical distributions
\end{itemize}

To encourage robustness, artificial masking is applied during training by randomly hiding observed values. The model is then trained to reconstruct these values:
\begin{itemize}
    \item Smooth L1 loss for numerical features
    \item Cross-entropy loss for categorical features
\end{itemize}

The contribution of the penalty and reconstruction terms is scaled dynamically based on prediction confidence. Confidence is estimated using the entropy of the predicted class probabilities, with lower entropy indicating higher confidence. The auxiliary losses are weighted proportionally to this confidence, encouraging stronger regularization when predictions are uncertain.

Training proceeds for a fixed number of iterations with mini-batch optimization. At each iteration:
\begin{itemize}
    \item A fraction of observed entries is randomly masked (data augmentation)
    \item Missing values are imputed using either the learned imputer or the reference baseline (with probability $0.3$)
    \item The classifier is trained on the resulting completed data
\end{itemize}

The Gumbel-Softmax temperature is annealed during training to encourage discrete categorical predictions. Early stopping is applied based on validation accuracy. At test time, missing values are imputed using the learned imputation network and predictions are produced by the classifier.

\paragraph{Number of parameters.}
Let $d$ denote the input dimensionality and $C$ the number of classes. The model consists of a classifier and a neural imputer. The total number of parameters is:
\[
\text{\# Parameters} = P_{\text{classifier}} + P_{\text{imputer}},
\]
where both terms depend only on $d$ and the network architecture, and are independent of the number of training samples.

\paragraph{Implementation details.} We use a hidden dimension of $h = 256$ for the imputation network.
Model parameters are initialized using Kaiming uniform for linear layers. Missing numerical entries are initialized using the reference values with additive Gaussian noise $\mathcal{N}(0, 0.01^2)$. For categorical features, missing values are initialized by sampling from a noisy version of the reference distribution via a softmax over perturbed log-probabilities. Observed categorical values are encoded as one-hot vectors.
The scaling factor for the residual correction is fixed to $\alpha = 0.1$. We use the Adam optimizer with learning rate $10^{-3}$ for the classifier and $5 \times 10^{-4}$ for the imputer. 
Training is performed for up to 300 iterations with mini-batches of size 64 and early stopping based on validation accuracy with a patience of $50$ iterations. Artificial masking is applied with probability proportional to the dataset missing rate, capped at $0.2$.
The Gumbel-Softmax temperature is initialized at $\tau = 1.0$ and annealed multiplicatively with a factor of $0.995$ per iteration down to a minimum of $0.3$. 
The penalty and reconstruction weights are set to $\lambda_{\text{pen}} = 0.1$ and $\lambda_{\text{rec}} = 0.1$. Confidence-based scaling is applied using normalized prediction entropy.

\paragraph{Computational Complexity.}
Let $n$ be the number of samples, $d$ the input dimensionality, $h$ the hidden dimension, $b$ the batch size, and $T$ the number of training iterations. The method jointly trains a neural imputer and a classifier, each requiring forward and backward passes with complexity $\mathcal{O}(dh + h^2)$ per sample.

The overall training complexity is:
\[
\mathcal{O}\left(T \cdot \frac{n}{b} \cdot (dh + h^2 + d)\right),
\]
where the additional $\mathcal{O}(d)$ term accounts for masking, reconstruction losses, and categorical sampling. The space complexity is $\mathcal{O}(nd + dh + h^2)$, dominated by storing the dataset, masks, and model parameters.

\section{Results}

\subsection{Overall stability}

These results are obtained through non-parametric factor effect analysis with effect sizes ($\eta^2$) to quantify the impact of dataset characteristics on downstream performance.

\paragraph{Joint-AmortizedImpute}

Table~\ref{tab:factor_impact} shows how dataset characteristics affect the relative performance of Joint-AmortizedImpute compared to the best classical method. The results indicate that feature dimensionality, dataset size,  missingness mechanism, number of classes, and imbalance, are statistically significant, though they present small effect sizes. In contrast, missing rate is not significant. Overall, the small effect sizes suggest that standard dataset characteristics explain only a limited portion of the performance differences, reinforcing that missingness structure, including intrinsic hardness, is the dominant factor. 

\begin{figure}[h!]
    \centering
    
    \begin{minipage}{0.48\linewidth}
        \centering
        \includegraphics[width=\linewidth]{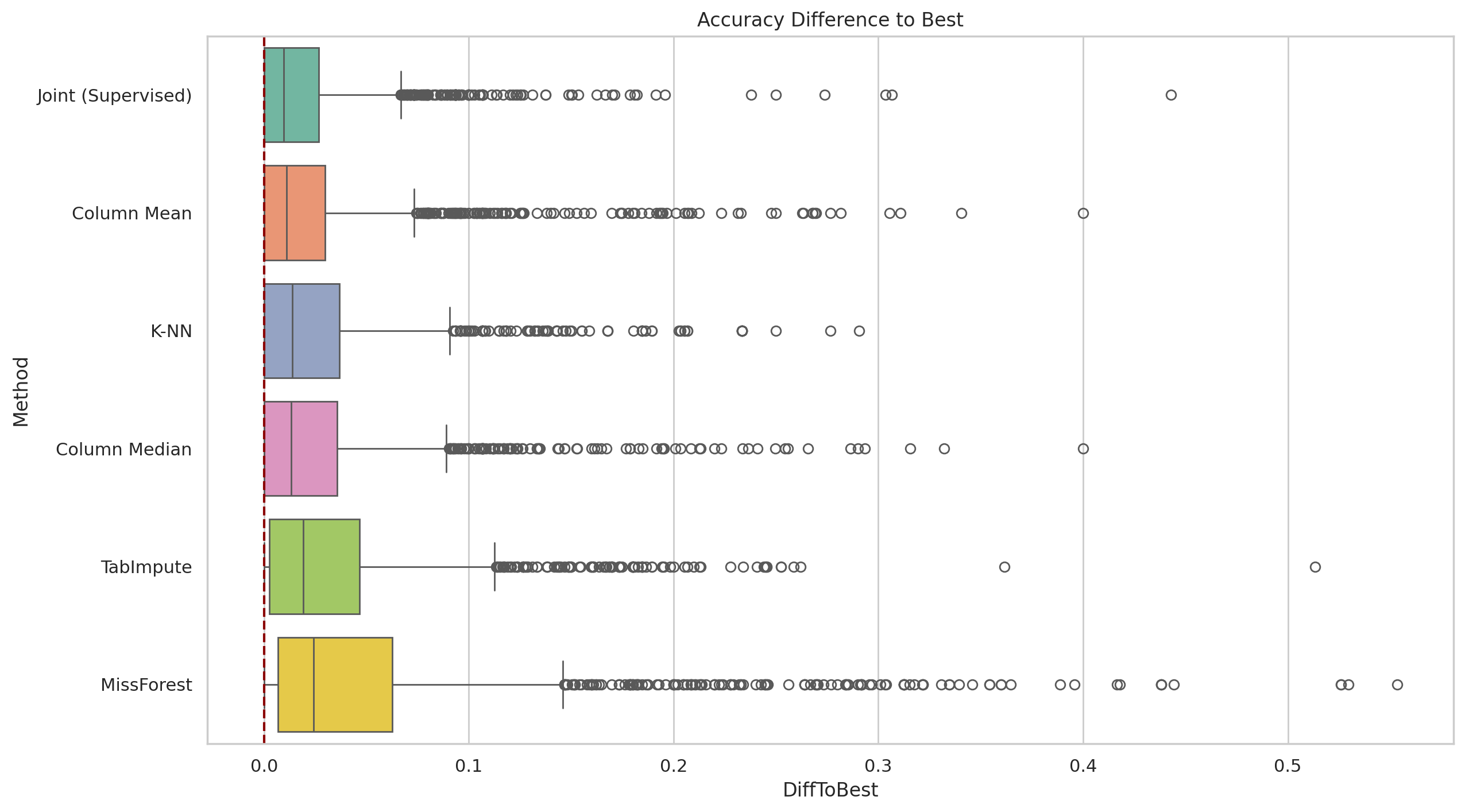}
        \caption{Accuracy difference to the best performing classical method on the test set.}
        \label{fig:acc_diff}
    \end{minipage}
    \hfill
    \begin{minipage}{0.48\linewidth}
        \centering
        \includegraphics[width=\linewidth]{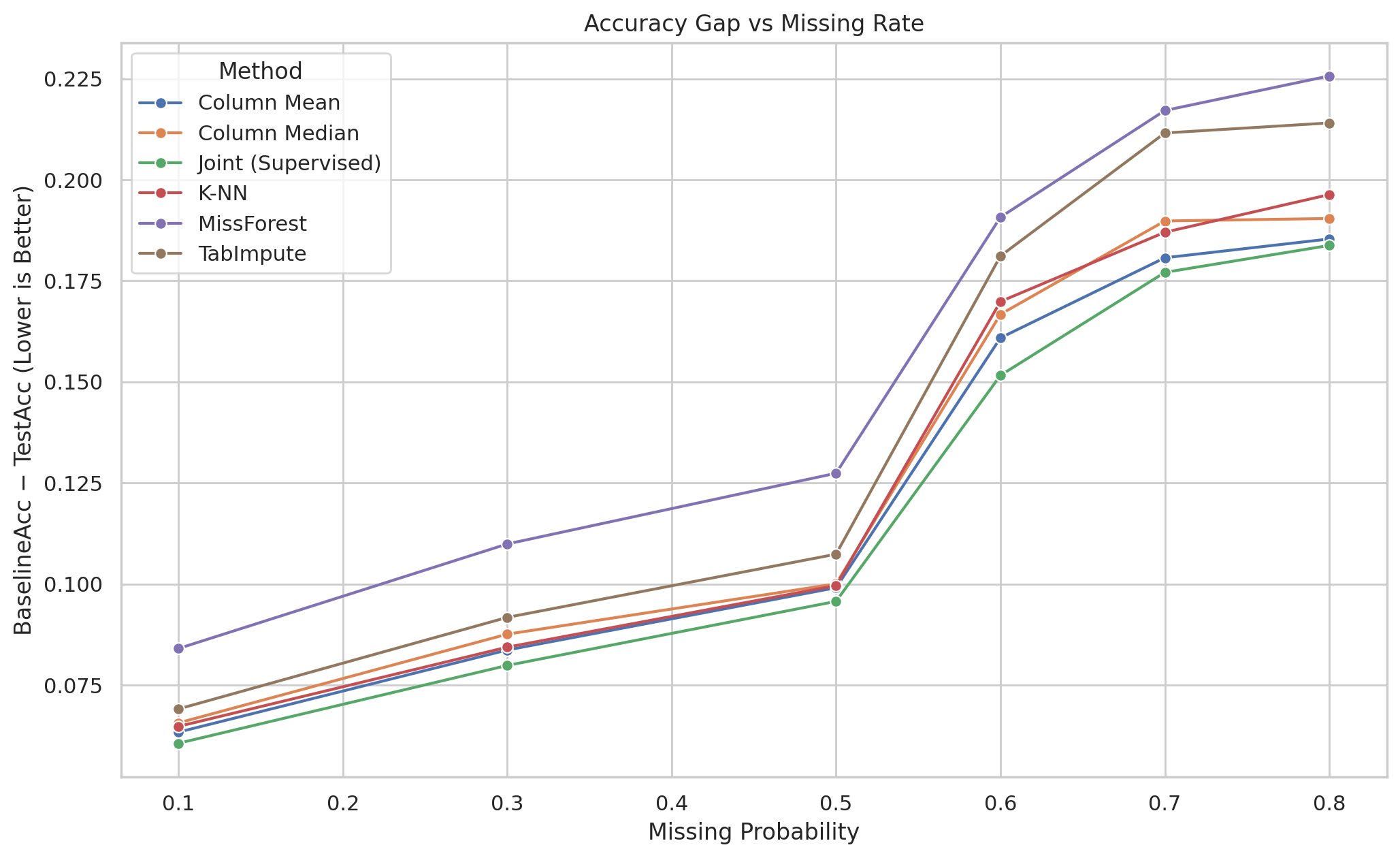}
        \caption{Accuracy gap to complete data baseline.}
        \label{fig:gap_missing}
    \end{minipage}
    
\end{figure}

Figure \ref{fig:acc_diff} shows the distribution of accuracy differences relative to the best-performing classical method on the test set across datasets. While methods often achieve similar performance (values close to zero), Joint-AmortizedImpute exhibits few extreme deviations, indicating stability.

Figure \ref{fig:gap_missing} shows the accuracy gap relative to the complete-data baseline as a function of the missing rate. As expected, performance degrades for all methods as missingness increases. However, Joint-AmortizedImpute consistently exhibits the smallest gap, indicating greater robustness to increasing levels of missingness. 

\begin{table}[h!]
    \centering
            \begin{tabular}{lcc}
        \toprule
        \textbf{Factor} & \textbf{Sig.} & $\boldsymbol{\eta^2}$ \\
        \midrule
        Imbalance      & \underline{Yes} & 0.005 \\
        Missing Rate   & No              & 0.0005 \\
        Dataset Size   & \underline{Yes} & 0.05 \\
        \#Categorical  & \underline{Yes} & 0.017 \\
        Mechanism      & \underline{Yes} & 0.015 \\
        \#Classes      & \underline{Yes} & 0.01 \\
        \bottomrule
        \end{tabular}
        \captionof{table}{Factor impact on performance differences between Joint-AmortizedImpute and the best classical method.}
        \label{tab:factor_impact}
\end{table}

\paragraph{Joint-LatentImpute.}

Overall, additional results on Joint-LatentImpute tell a similar story.

Table~\ref{tab:factor_analysis_direction_m1} shows that the number of categorical features, dataset size and missingness mechanism have a statistically significant effect on the performance difference between the proposed Joint-LatentImpute and the best benchmark imputation method. Imbalance, missing rate and number of classes do not show a statistically significant effect. Once again, the small effect sizes suggest that differences in performance are not fully explained by dataset properties.

\begin{figure}[h!]
    \centering

    \begin{minipage}{0.48\linewidth}
        \centering
        \includegraphics[width=\linewidth]{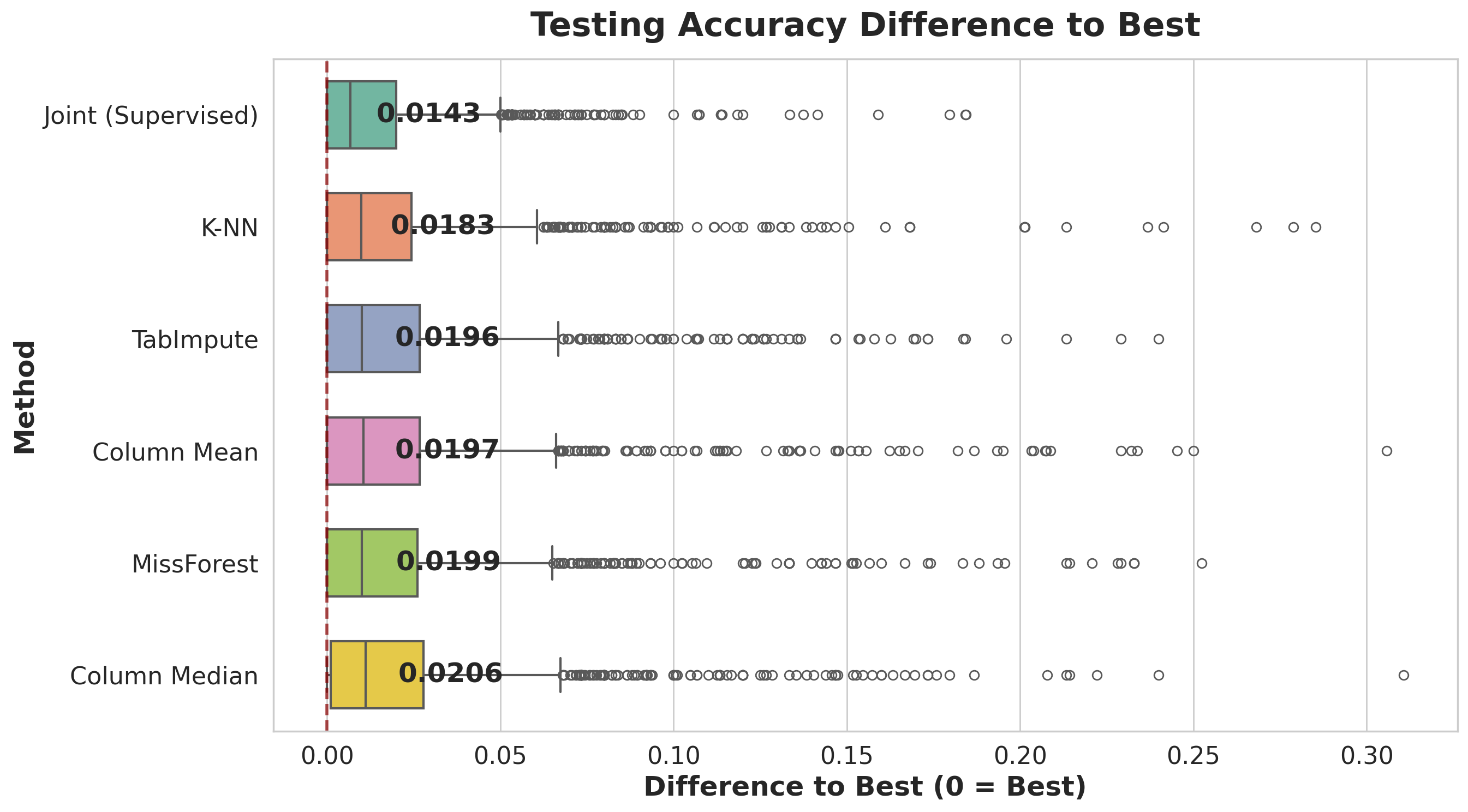}
        \caption{Accuracy difference to the best classical method.}
        \label{fig:acc_diff_jli}
    \end{minipage}
    \hfill
    \begin{minipage}{0.48\linewidth}
        \centering
        \includegraphics[width=\linewidth]{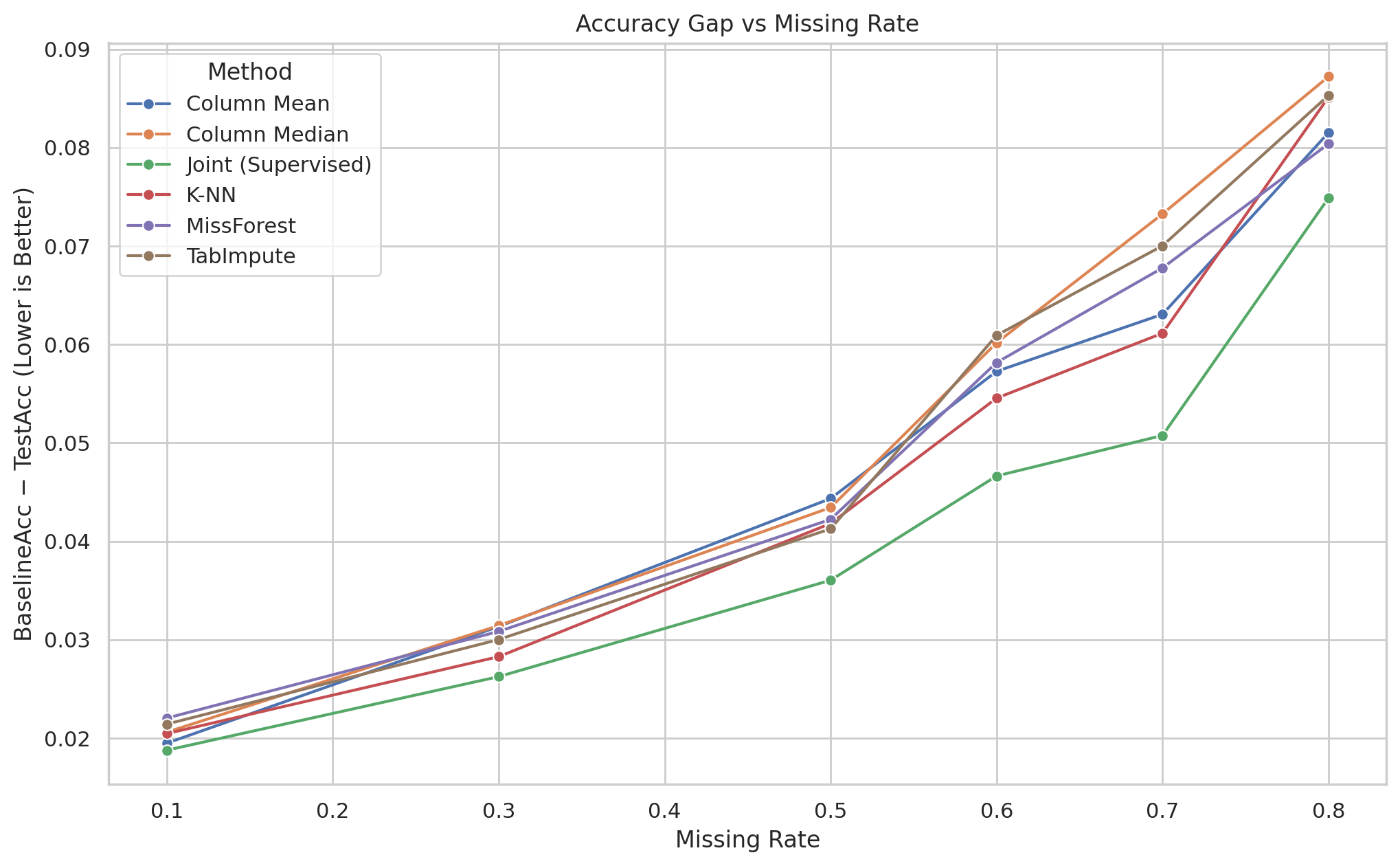}
        \caption{Accuracy gap vs missing rate (all methods).}
        \label{fig:gap_missing_jli}
    \end{minipage}

\end{figure}

Figure~\ref{fig:acc_diff_jli} shows the distribution of accuracy differences relative to the best-performing classical method on the test set across datasets. Joint-LatentImpute remains consistently close to the best method, with a tighter distribution and fewer extreme degradations, suggesting improved robustness compared to classical approaches.

Figure~\ref{fig:gap_missing_jli} shows the accuracy gap relative to the complete-data baseline as a function of the missing rate. Joint-LatentImpute consistently exhibits the smallest gap, indicating greater robustness to increasing levels of missingness.

\begin{table}[h!]
    \centering
            \begin{tabular}{lcc}
        \toprule
        \textbf{Factor} & \textbf{Sig.} & $\eta^2$ \\
        \midrule
        Imbalance      & No  & 0.002 \\
        Missing Rate   & No  & 0.0002 \\
        Dataset Size   & Yes & 0.03 \\
        \#Categorical  & Yes & 0.04 \\
        Mechanism      & Yes & 0.02 \\
        \#Classes      & No  & 0.003 \\
        \bottomrule
        \end{tabular}
        \captionof{table}{Factor impact on performance differences between Joint-LatentImpute and the best classical method.}
        \label{tab:factor_analysis_direction_m1}
\end{table}

\subsection{Computational cost}

\paragraph{Joint-AmortizedImpute.} The proposed Joint-AmortizedImpute introduces an upfront training cost, with a mean training time of approximately 39.3 seconds due to the joint optimization of the imputation and predictive components. 
However, once trained, imputation is performed via a single forward pass, resulting in efficient inference. 

\paragraph{Joint-LatentImpute.}
We compare the computational cost of the proposed method with classical imputation approaches. 
As expected, simple methods such as mean and median imputation incur negligible cost, while more advanced methods such as MissForest and TabImpute require several seconds per dataset.
The proposed joint method has a mean imputation time of approximately 31.7 seconds due to the end-to-end optimization of imputations and model parameters. 
We report wall-clock time required to impute the training data. For iterative methods such as MissForest and TabImpute it includes model fitting.

\begin{table}[h!]
    \centering
    \begin{tabular}{lrrr}
\toprule
Method & mean ($\downarrow$) & median ($\downarrow$) & std ($\downarrow$) \\
\midrule
Column Mean & \textbf{0.0009} &\textbf{ 0.000 }& \textbf{0.002 }\\
Column Median &\textbf{ 0.0009 }& \textbf{0.000} & 0.003 \\
K-NN & 0.692 & 0.195 & 1.4 \\
MissForest & 12.02 & 8.51 & 11.23 \\
TabImpute & 21.94 & 7.15 & 57.47 \\
Joint-LatentImpute (Ours) & 31.68 & 12.66 & 48.75 \\
\bottomrule
\end{tabular}
    \caption{Average time (in seconds) required to obtain a fully imputed training dataset across all datasets. 
    For iterative methods (MissForest and TabImpute), this includes model fitting.}
    \label{tab:placeholder}
\end{table}

\end{document}